\documentclass{article}

\usepackage{tabu}
\usepackage{booktabs}% for better rules in the table
\usepackage{times}
\usepackage{graphicx} % more modern
\usepackage{tabularx}
\usepackage{subfigure} 
\usepackage{amsmath}

\usepackage{algorithm}
\usepackage{algorithmic}

\usepackage{hyperref}

\usepackage[accepted]{icml2017}

\icmltitlerunning{Does Your Phone Know Your Touch?}

\begin{document} 

\twocolumn[
\icmltitle{Does Your Phone Know Your Touch?}
\title{Does Your Phone Know Your Touch?}

% It is OKAY to include author information, even for blind
% submissions: the style file will automatically remove it for you
% unless you've provided the [accepted] option to the icml2017
% package.

% list of affiliations. the first argument should be a (short)
% identifier you will use later to specify author affiliations
% Academic affiliations should list Department, University, City, Region, Country
% Industry affiliations should list Company, City, Region, Country

% you can specify symbols, otherwise they are numbered in order
% ideally, you should not use this facility. affiliations will be numbered
% in order of appearance and this is the preferred way.
\icmlsetsymbol{equal}{*}

\begin{icmlauthorlist}
\icmlauthor{John Peruzzi}{stanford}
\icmlauthor{Philip Andrew Wingard}{stanford}
\icmlauthor{David Zucker}{stanford}
\end{icmlauthorlist}

\icmlaffiliation{stanford}{Stanford University Stanford, California}

\icmlcorrespondingauthor{Cieua Vvvvv}{c.vvvvv@googol.com}
\icmlcorrespondingauthor{Eee Pppp}{ep@eden.co.uk}

% You may provide any keywords that you 
% find helpful for describing your paper; these are used to populate 
% the "keywords" metadata in the PDF but will not be shown in the document
\icmlkeywords{boring formatting information, machine learning, ICML}

\vskip 0.3in
]

% this must go after the closing bracket ] following \twocolumn[ ...

% This command actually creates the footnote in the first column
% listing the affiliations and the copyright notice.
% The command takes one argument, which is text to display at the start of the footnote.
% The \icmlEqualContribution command is standard text for equal contribution.
% Remove it (just {}) if you do not need this facility.

%\printAffiliationsAndNotice{}  % leave blank if no need to mention equal contribution
%\printAffiliationsAndNotice{\icmlEqualContribution} % otherwise use the standard text.

%Abstract
%\begin{abstract} 
%With the demand for a secure personal cell phone while simultaneously unlocking the device in a natural and effortless manner, this paper looks into touchscreen based classification of a user and rejection of authentication from an unknown person. Although smart phone users may all be capable of tapping or drawing a specific character on a device, we propose that unique behavioral patterns can be extracted from a person to classify who specifically is using the device. 
%The problem is handled as both a classification problem from a set of known users and an unsupervised problem in learning a single user's behavior to reject all anomalies during touch collection from the device. 
%\end{abstract} 

% Intro
\section{Introduction}
\label{introduction}

In this paper, we consider the problem of distinguishing between authorized and unauthorized touchscreen and smart phone device users by leveraging a learned gesture classification profile combined with gesture anomaly detection.\newline
As touchscreen devices become more ubiquitous and the information stored within them becomes increasingly personal and valuable, the incentive and reward for circumventing existing security mechanisms has increased substantially.  Given known vulnerabilities in existing biometric and non-biometric authentication methods such as fingerprint scanners, facial recognition, tokens, and pass codes; the development of an effective authentication approach that goes beyond the 'something you know / have / are' paradigm is needed.  In this paper, we propose models that accurately predict users based on touchscreen gesture patterns and detect anomalies in these patterns as a versatile approach to augment existing security methods and provide a method of continuous authentication.  \newline
Touchscreen gesture are collected from a set of users from a capacitive sensor array to simulate a smart phone. Features include the pressure measured at the two dimensional (X,Y) coordinates on the sensor for each gesture, velocity at different instances of the gesture, and the duration of the gesture.  We then demonstrate how logistic regression, support vector machines (SVM), and multiple Gaussian processes can be used to classify and predict the user creating the gesture.  \newline
Our intent is to determine the extent to which supervised and unsupervised learning approaches can be successfully leveraged across multiple domains to limit the impact of unauthorized touchscreen device usage, quantify touchscreen security weaknesses and vulnerabilities, and potentially inform touchscreen device security design.  Scenarios where our analysis may be useful include high security use cases where continuous authentication is required in the finance, transportation, public safety sectors.

\section{Related Work} 
Smart phone user authentication through biometric data has become an area of substantial interest. Recent literature exists describing attempts to classify smart phone users in a binary class supervised environment where a single user is predicted among many anonymous users. These projects have found success using kNN (with $k=1$). A combined effort from the University of Houston and Samsung Research America achieving accuracy of 90\% [1].
Neural networks have also been successful in classify users with accuracy up to 90\%, as demonstrated by a research group from the City University of Hong Kong [2]. Other groups, such as one at Arizona State University, found that SVMs can classify users with 90\% accuracy [3].  Other projects have also attempted one-class models. One-class SVMs have been shown to have incredible success rates, with one project from the College of William and Mary reporting accuracy over 97\% [4].  It is difficult to directly compare our accuracies to these projects. In most cases, while our accuracies were higher, the lack of publicly available datasets meant we had to conduct our own data collection, and thus we suffered from a lack of anonymous data to compare against, making our classification tasks simpler. \newline
We attempted a mixture of Gaussian models instead, and achieved accuracy comparable to the College of William and Mary project. Another report, tested both two-class and one-class algorithms using an SVM for both models [5]. This report found poor performance when trained solely on touch gestures, and therefore used other data from the gyroscope and more macroscopic movements combined with their touch data to achieve impressive results.\newline
While previous work has achieved relative success with different algorithms, our project is unique in that it compares one-class and two-class techniques. We also achieved the interesting preliminary result that the often used SVM two-class identifier seemed less versatile than expected and, in some cases, performed worse than simple logistic regression, which many previous studies did not discuss. \newline
In contrast to other projects, the scope of our data is limited.  Other projects have collected data from 20-100+ users [2][4][1]. Regarding features, we tended to make similar choices as previous work, prioritizing pressure, duration, and speed metrics [2].

% Data Acquisition and Features 
\section{Data Acquisition and Feature Selection} 
Modern smart phones leverage a capacitive touchscreen to capture finger movement and determine gestures. To simulate user gesture data, a capacitive sensor array was used under the assumption that the data presented will be similar to the internal data collected on a smart phone during typical use.
\subsection{Raw Data}
In this paper, we collected data using a capacitive sensor array as described above from a set of 5 individual users.  The sensor collected data at 30 frames per second with each frame representing a two dimensional array (corresponding to the X,Y plane) and corresponding to the  pixels on the sensor. Each value measured as the change in capacitance, due to the presence of a finger or other conductive object. Therefore, pixel values in each reported frame are  proportional to the finger area covering the selected pixels, inversely proportional to the distance from the sensor, and inherent random noise.

\begin{table}
\label{data-table}
\vskip 0.15in
\begin{center}
\begin{small}
\begin{sc}
\begin{tabularx}{\columnwidth}{Xccc}
\hline
\abovespace
\belowspace
Dataset & Taps & Circles & Random \\
\hline
\abovespace
Samples (n)			& 1515 & 557 & 271 \\
Classes   			& 4 & 4 & 2 \\
Coupling (Pressure) \\
features per frame	& 25 & 49 & 49 \\
Velocity \\	
features per frame  & 0 & 2 & 2 \\
Frames   			& 5 & 30 & 30 \\
Duration Features		& 1 & 1 & 1 \\
\hline
\abovespace
Total Features		& 126 & 1569 & 1569 \\
\hline
\end{tabularx}
\end{sc}
\end{small}
\end{center}
\vskip -0.2in
\caption{Characteristics of datasets included in this study.}
\end{table}
\subsection{Development of Feature Sets}
Data was collected from a group of five different individuals who were asked to repetitively perform three gestures with the capacitive sensor as they would on their phone - tap, draw a circle, and randomly draw different patterns. Each of these three gesture categories were  isolated and a divided into discrete events.  An event was defined as beginning when a finger first touches the sensor to when the finger stops touching the sensor. A finger is considered to be on the panel if the maximum value of a frame exceeds a set threshold. These events are recorded for processing as follows. First, we specify the size of a window to use when examining the gesture. The center of the finger is located by the sensor maximum value, then a square window is cut around the finger to be used as features. Window sizes were $nxn$ pixels with $n \in [3,5,7]$, resulting in frame sizes of 9, 25, and 49, respectively. Next we specified the number of time sequence frames, $f$ to capture during the gesture with $f \in (3,5,30)$. If the gesture consisted of additional frames, we selected a linearly separated set of frames to represent the gesture. If there were fewer frames in the gesture the required, we captured the windows around the gesture and interpolated the shape of the finger centered around the maximum. Additional features were tested along with the finger pressure at different pixels, including duration of a gesture, and velocity of finger in the x-axis and y-axis. 

\begin{figure}
\begin{center}
\includegraphics[width=8cm]{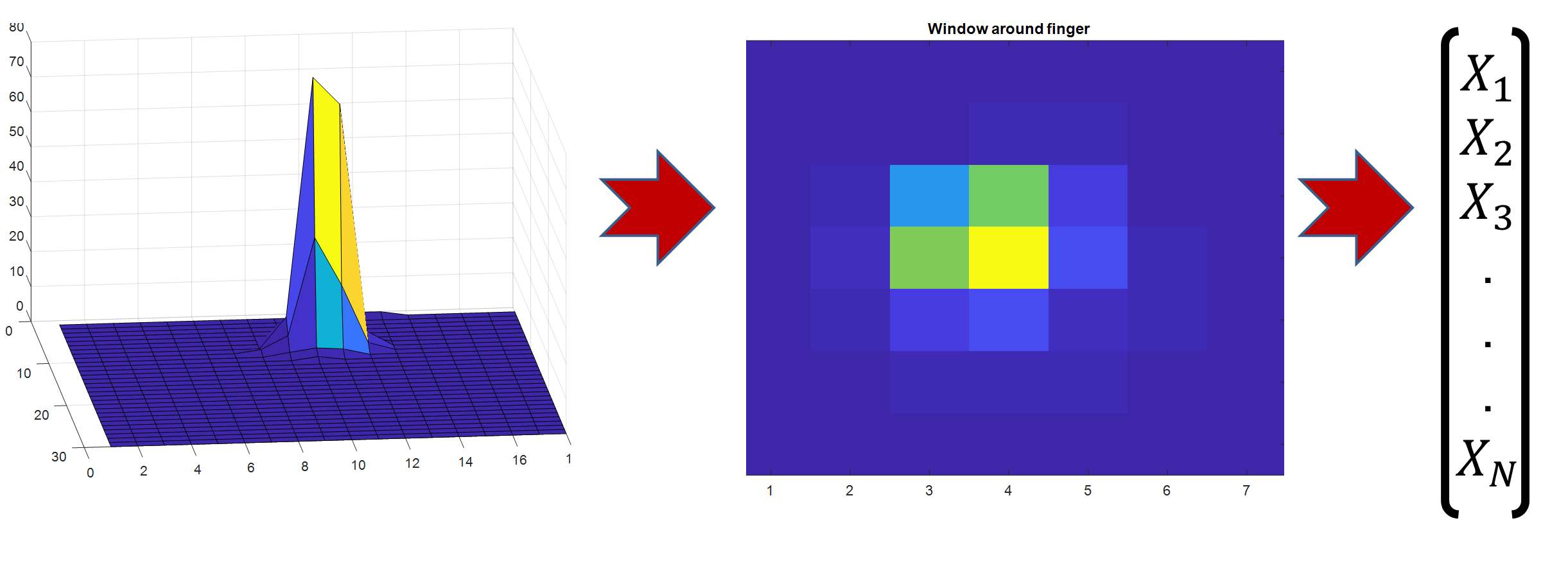}
\caption{Conversion from frames to feature sets}
\end{center}
\end{figure}

\subsection{Feature Selection and Dimensionality Reduction}
Our datasets are characterized as having a large number of features as compared to the number of samples.  With the exception of Taps, $n << m$.  To help address this problem, we implemented to feature selection and dimensionality reduction approaches - recursive feature estimation (RFECV) and PCA.  

To determine the most important modeling features in each gesture dataset, we executed RFECV, a backward-based recursive process [6] with $k=10$ cross validation.  RFECV builds an initial model with all available features and iteratively removes features until the best set of model features is identified as evaluated by the predictive fit an SVM classifier using a linear kernel.

The RFECV process yielded 15 features for the Taps dataset, 11 features for the Circles dataset, and 1 feature for the Random dataset.  When these features were overlayed onto a single (X,Y) frame, the following patterns emerged and are illustrated in Figure 2. 

\textbf{Taps} - The most important features for Taps are the origin of the frame and tap duration. 
An additional four of eight points surrounding the origin were identified as significant. The most predictive information about a tap is the points surrounding the origin and the tap duration. 
\newline
\textbf{Circles} - The most important features for Circles are the one point towards the upper right of the frame and gesture duration. An additional eight points within the frame identified as significant. The significance of the single point may be pressure exerted during the beginning or end of a circle is indicative of the user.
\newline
\textbf{Random} - The most important feature for Random was the upper-right most point in the frame.  Since there were only two user classes for Random data, this likely indicates one of the two users touched the edge of the frame enough times to sufficiently predict the correct user class.  
%Since the RFECV algorithm identifies the smallest set of features, the result is the one predictive feature.  Stopping RFECV sooner or increasing the data quality / diversity may mitigate this issue. 
\newline
Other important findings include gesture duration is only significant for taps and circles,  and the size of the frame (5x5 vs 7x7) is not significant since many of the points on the frame edge were not chosen as predictive.\newline
PCA was successful in reducing the dimensionality of the data.  Specifically, PCA was able to reduce Taps data into 20 dimensions while accounting for 90\%+ of the original dataset variance.  However, PCA was not as successful with the Circles or Random datasets.  Given the success of RFECV across datasets and partial success of PCA, we proceeded with RFECV.
%\begin{itemize}
%\item Include feature selection methodology (RFECV) and PCA, comparison of important variables / dimensions.
%\item Add backward selection algorithm.  
%\item Consider 2x2 plot of PCA1 vs PCA2 for Taps
%\item zero centered , standardized data
%\item randomized, standard pickled datasets 
%\item list of chosen features
%\item finding  - only one feature chosen for random
%\item finding - frame size not important, speed not important
%\item figures - heatmap of chosen features, possible PCA
%\end{itemize}

\begin{figure}
\begin{center}
\includegraphics[width=\columnwidth]{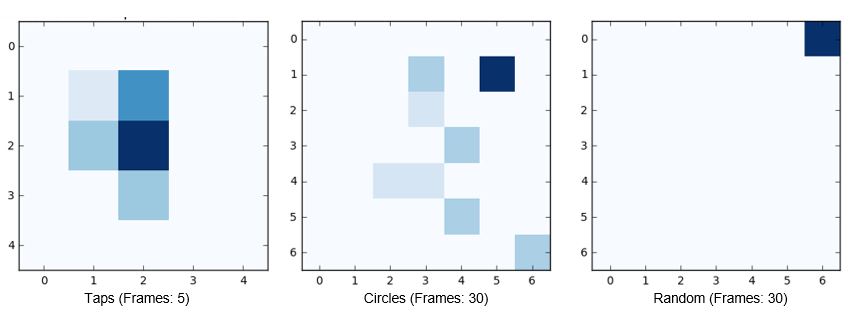}
\caption{Heatmap of (X,Y) coordinate features per dataset post-feature selection}
\end{center}
\end{figure}

%\begin{figure}[H]
%\begin{center}
%\includegraphics[width=8cm]{PCA_With_CDF.jpg}
%\caption{Effect of capturing variation using PCA}
%\end{center}
%\end{figure}

%\begin{figure}[ht]
%\vskip 0.2in
%\begin{center}
%\centerline{\includegraphics[width=3cm]{Taps_5_PCA}}
%\caption{Caption Text}
%\end{center}
%\vskip -0.2in
%\end{figure} 

% Learning Methods
\section{Learning Methods}
%We first implemented algorithms for the simpler task of supervised classification trained on two or more classes and later attempted the more difficult of anomaly detection training only one class. 
%We use logistic and softmax regression for classification of users. For anomaly detection, we train on data from only one user using a multiple Gaussian algorithm.
%, and then attempt to classify test gestures from multiple users as either normal or anomalous. 

\subsection{Logistic and Softmax Regression}

Logistic regression was chosen as our first model due to its simplicity and intended use as a classifier. We used logistic regression to distinguish between two users we had already seen in training. The logistic hypothesis for binary classification for a given training example x is as follows:
\begin{equation}
	h_{\theta}(x) = \frac{1}{1+e^{-\theta^{T}x}}
\end{equation}
The goal is to learn the the most likely parameters, $\theta$, given our training data.%given by choosing the $\theta$ which maximizes the following equation over the training set of m examples:
%\begin{equation}
	%\sum_{i=1}^m y^{(i)} log(h(x^{(i)})) + (1 - y^{(i)}) log(1 - h(x^{(i)}))
%\end{equation}
  We also applied softmax regression to see if we could differentiate all four users in a supervised setting. The softmax hypothesis outputs a vector of probabilities for each of the k possible output classes, with the probability for class $c$ for a given input $x$ being:
\begin{equation}
	\frac{exp(\theta_{c}^{T}x)}{\sum_{j=1}^k exp(\theta_{j}^{T}x)}
\end{equation}
Additionally, we experimented with regularization to discourage over-fitting. 
%However, due to our high model accuracy, we found it difficult to identify optimized hyper-parameters as there was not much that could be done to improve our model on our dataset.

\subsection{Support Vector Machines (SVM)}
The support vector machine algorithm was chosen as our second algorithm for its known supervised classification capabilities.  We initially experimented with a binary classifier and expanded to a multiclass classifier.  The objective of an SVM is to identify a separating hyperplane that separates the data into discrete classes within a multidimensional space.  The SVM classifier and functional margin are written, respectively, as the following:
\begin{equation}
\begin{aligned}
	& h_{w,b}(x) = g(w^Tx+b) \\
	& \hat{\gamma}^{(i)} = y^{(i)}(w^Tx+b)
\end{aligned}
\end{equation}
Given a training dataset $S = \{(x^{(i)}, y^{(i)}); i = 1, ..., m\}$, we want to find the optimal margin classifier using the optimization below:
\begin{equation}
\begin{aligned}
	& min_{\gamma,w,b} \frac{1}{2}||w||^2 \\
   	& s.t.\space \space y^{(i)}(w^T x+b) >= 1; i = 1, ..., m    
\end{aligned}
\end{equation}
We experimented with various SVM kernels including the linear, polynomial, and radial basis function (RBF) kernels and determined RBF most closely met our needs. 

\subsection{Multiple Gaussian Anomaly Detection}
This method learns the intricacy of the user's finger shape and movement. The method of Multiple Gaussian Anomaly detection uses an iterative Expectation-Maximization algorithm on multiple Gaussians to find $k$ different multinomial distributions. 
Typically when fitting a Gaussian distribution, the log probability of the pdf is maximized.  However, we add the latent random variable $z$ where \(	{ z^{(i)} \sim Multinomial (\phi) (Where, \phi_j \ge 0 , \sum_{j = 1}^k \phi_j = 1)}\). So we have the expression below to maximize. 
\begin{equation}
\begin{aligned}
	& l( \phi, \mu , \Sigma)  = \sum_{i = 1}^m log \, 	p(x^{(i)};\phi, \mu, \sigma) \\ 
    &	 \qquad =  \sum_{i = 1}^m log \sum_{z^{(i)}}^k  p(x^{(i)}|z^{(i)};\mu, \sigma) p(z^{(i)};\phi) \\ 
\end{aligned}
\end{equation}
Because $z$ is an unknown latent variable, we are unable to solve for the above expression directly and use the iterative EM algorithm. 
\begin{equation}
\begin{aligned}
	& Repeat\ until\ convergence:  \\
	& \quad (E-step) \ For \ each \ i,j, \ set  \\
    & \qquad Q_j^i := p(z^i = j | x^i; \phi, \mu, \Sigma ) \\ 
    & \quad (M-step) \ Update \ the \ parameters: \\
    & \qquad \phi_j := \frac{1}{m} \sum_{i = 1}^m Q_j^i, \\
    & \qquad \mu_j := \frac{\sum_{i = 1}^m w_j^i x^i }{\sum_{i = 1}^m w_j^i}, \\
    & \qquad \Sigma_j := \frac{\sum_{i = 1}^m w_j^i (x^i - \mu_j)(x^i - \mu_j)^T }{\sum_{i = 1}^m w_j^i} \\
\end{aligned}
\end{equation}

After converging the parameters above and solving for the multiple Gaussian, we can select a threshold to be considered the user. If the maximum probability that a gesture belongs to one of these clusters is below the selected 'user' threshold, the gesture is considered to not be the user and is rejected. 

% Experiement Results
\section{Experiment Results}
\begin{figure}
\begin{center}
\includegraphics[width=4cm]{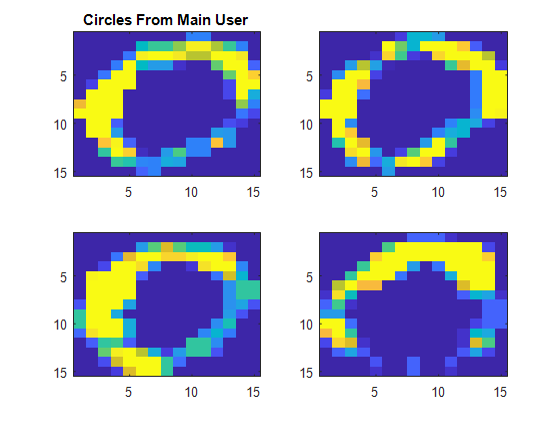}
  \includegraphics[width=4cm]{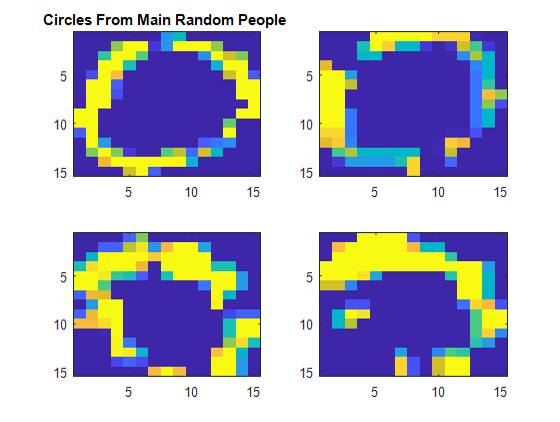}
\caption{Examples of separate users drawing circles.}
\end{center}
\end{figure}

\subsection{Logistic and Softmax Regression}
 Below are our confusion matrices for logistic and softmax regression. Our test accuracies ranged from 95\% to 100\% for all gestures and all  user combinations.
\begin{figure}[H]
\begin{center}
\includegraphics[width=8.2cm]{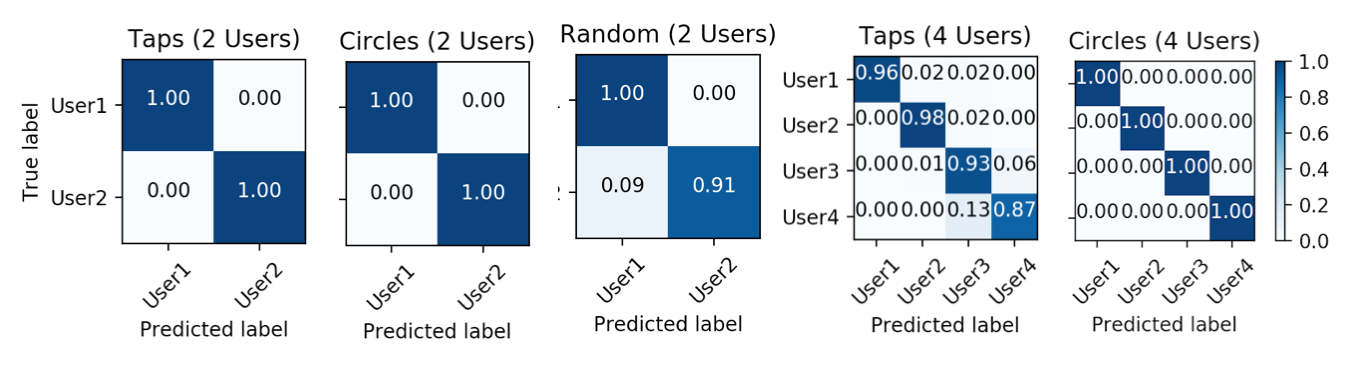}
\caption{Confusion Matrices for binary and softmax regression and various gestures. Note: only two users gave random gestures.}
\end{center}
\end{figure} 
\vskip -0.2in

Even though our initial feature analysis proved that only a few principal components or features are necessary to account for variation in the data, logistic accuracies, unlike those with our other models, were generally not impacted by feature size, with larger dimensional data often yielding superior results. Final logistic results are reported with the maximum feature amount and no regularization in Table 3. %An example of this can be seen in Figure 5.

%\begin{figure}
%\begin{center}
%\includegraphics[width=4.5cm]{Screen_Shot_2017-12-14_at_4_08_36_PM.png}
%\caption{Example plot of logistic regression accuracies at various feature dimensions for binary, circle classification.}
%\end{center}
%\end{figure} 

\subsection{Support Vector Machines (SVM)}
%... to do:
%\begin{enumerate}
%\item create AUC/ROC chart
%\item fix chart headers, figure references
%\end{enumerate}

We implemented SVM with the RBF kernel on the Taps, Circle and Random gesture datasets leveraging the set of features identified by the recursive feature estimation process.  To determine the best SVM model parameters, we ran a grid search for C and $\gamma$ between (1e-03 thru 1e+10) and (1e-15 thru 1e+03), respectively.  The data was split with 80\% of the data allocated to training and 20\% to testing.\newline
SVM classification worked very well in predicting Users for the Taps and Random datasets, however, not as well for the Circle dataset.  With tuned parameters, SVM was able to correctly predict classifications on 96\% of Taps samples, 70\% of Circles samples, and 98\% of Random samples.  Notably, prior to running recursive feature estimation, SVM correctly predicted classifications on 51\% of the Random data as all data points for User 2 were misclassified.  Similar performance was noted with Gaussian anomaly detection. \newline
%\textcolor{red}{Is this true?} yes it is how I got improved taps performance. 
Important findings include $C$ and $\gamma$ are large for Circles and Random datasets, respectively.  This indicates the model is allowing a large boundary or accepting a high cost in order to classify these data points. \newline
Details for each dataset's classifier can be found in Table 2 and parameter tuning results can be found in Figure 6.  Confusion matrices with can be found in Figure 7. \newline

\begin{table}
\vskip -0.2in
\caption{Characteristics of optimal SVM Models.}
\label{SVM-model-table}
\vskip 0.15in
\begin{center}
\begin{small}
%\begin{sc}
\begin{tabularx}{\columnwidth}{Xlll}
\hline
\abovespace
\belowspace
Characteristics & Taps & Circles & Random \\
\hline
\abovespace
Kernel			& RBF & RBF & RBF \\
C				& 10.0 & 1000.0 & 0.10 \\
$\gamma$		& 0.01 & 0.01 & 1000.0 \\
F1-Score   		& 0.96 & 0.69 & 0.98 \\
%\hline
%\abovespace
%Test Accuracy		& 96.0\% & 70.0\% & 98.0\% \\
%Train Accuracy   	& 100.0\% & 96.0\% & 98.0\% \\
\hline
\end{tabularx}
%\end{sc}
\end{small}
\end{center}
\vskip -0.2in
\end{table}

\begin{figure}
%\vskip -0.1in
\begin{center}
\includegraphics[width=7cm]{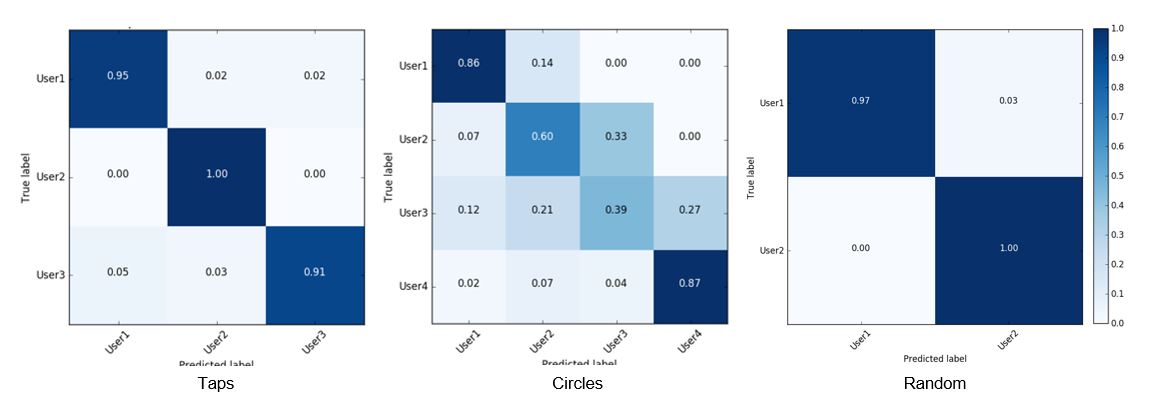}
\caption{SVM Confusion matrices for Taps, Circles and Random datasets.}
\end{center}
\vskip -0.1in
\end{figure} 
\vskip -0.1in

% do we want these svm charts across the page? seem to take up a decent amount of room?

\begin{figure*}
\vskip -0.1in
\begin{center}
\begin{minipage}[c]{0.3\linewidth} 
\includegraphics[width=\linewidth]{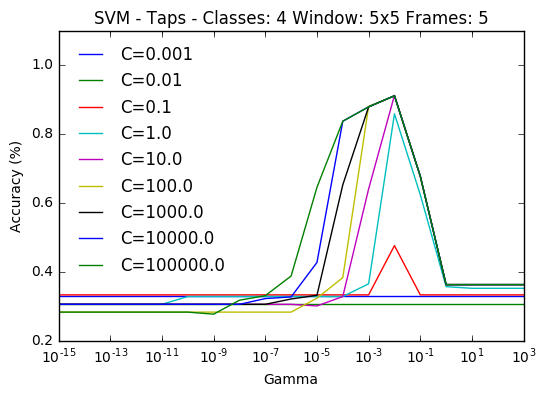}
\end{minipage}
\hfill
\begin{minipage}[c]{0.3\linewidth}
\includegraphics[width=\linewidth]{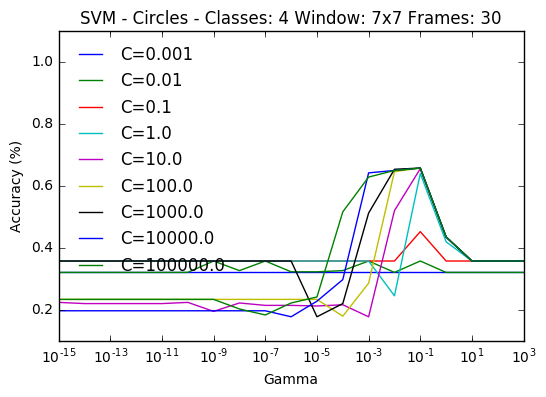}
\end{minipage}%
\hfill
\begin{minipage}[c]{0.3\linewidth}
\includegraphics[width=\linewidth]{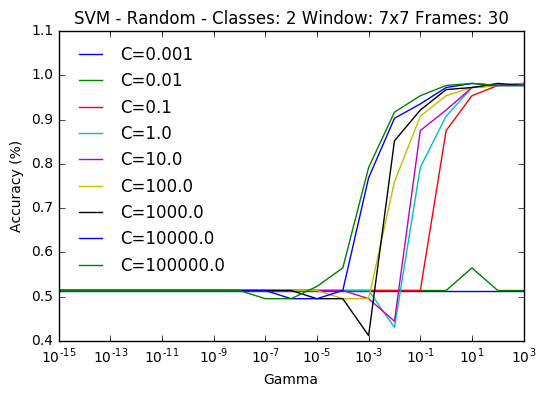}
\end{minipage}%
\vskip -0.1in
\caption{SVM classification results across datasets using parameter grid search for C and $\gamma$ between (1e-03 thru 1e+10) and (1e-15 thru 1e+03), respectively.}
\end{center}
\vskip -0.1in
\end{figure*}

\subsection{Multiple Gaussians Anomaly Detection}
Performance for Taps and Circles averaged over 90\({\%}\). \newline The Random data had the lowest performance with a true rejection rate of 78\({\%}\). On datasets with a large number of features, the accuracy between train and test data varied widely without regularization. Typically, this implies a variance problem and the algorithm is over fitting the training set.  When feature selection and data normalizing was introduced, Taps performance jumped from 78\({\%}\) to 92\({\%}\). Anomaly detection results can be found in Table 3. The regularization option for Multiple Gaussian adds a small value to the diagonal components in the covariance matrix. This essentially increases the variance of each variable, as the diagonal components of a covariance matrix is the measured variance for each respective variable.

%Ultimately performance for taps and Circles did well with  average perforamce above 90\({\%}\). Random movements typically had the worst performance with a correct rejection rate of 78\({\%}\). Although performance may be correlated to smaller training set and thus the requirement for a smaller feature set. 
%\newline
%There was a large separation between training set accuracy and testing set accuracy on data with large feature sets, normalization had to be used with a value above 1. Typically a separation of training and testing set implies we have a variance problem, and the algorithm is over fitting. I believe some of the pixels that have low variance are providing little information to the model. Conceptually, some of the far pixels in a frame may stay constant as a finger is never large enough to alter the value. When using feature selection and normalizing the data, taps performance had jumped from 78\({\%}\) to 92\({\%}\). 
% * <scandreww@gmail.com> 2017-12-16T00:00:29.985Z:
%
% ^.

%\begin{figure}
%\begin{center}
%\includegraphics[width=\columnwidth]{VsRegularization.jpg}
%\caption{Multiple Gaussian on circles has dramatic improvement sweeping over regularization.}
%\end{center}
%\end{figure} 

\subsection{Summary, Analysis, and Additional Results}
Overall model results are summarized in Table 3.

Logistic regression was our best performing model with minimum accuracy of over 94.9\% across model variations, parameters, and datasets. Although some models performed better on specific datasets, these results suggest two-class learning with preloaded gesture data could provide a viable anomaly detection mechanism - potentially superior to the one-class anomaly detection mechanism often found in previous literature. Interestingly, when logistic regression was trained on one primary user and two unauthorized users, and tested on a never-before seen user, our model consistently rejected the unseen user and accepted a new instance of gestures from the primary user, often reaching accuracies of up to 93\%. \newline
%Additionally, some of our results are so accurate that they necessitate additional explanation. 
Data exploration reveals that our data is highly separable. This is likely due to an inorganic data collection process with excessively repetitive gestures. As evidence of this, when trained in a supervised setting, two data-collection instances from the same user at different times of the day could be differentiated with an accuracy over 90\% with logistic regression. 
\newline
Other results indicate our models do learn attributes inherent to the user and not just the training instance. Our models had more difficulty differentiating between two time-separated instances from the same user than between that user and a different user, indicating that we actually learned traits inherent to the user. Additionally, our Gaussian mixture model improved its accuracy when trained and tested on two time separated instances of a user in comparison to training and testing on only a single instance. This demonstrates that more varied, realistic, and organic data from a single user could actually still yield impressive results, as it makes the Gaussian less rigid.
\begin{table}
\caption{Classification accuracies for Logistic Regression, Support Vector Machines, and Multiple Gaussians across datasets.}

\label{results-table}
\vskip -0.15in
\begin{center}
\begin{small}
\begin{sc}
\begin{tabular}{lccr}
\hline
\abovespace
\belowspace
Model & Taps & Circles & Random \\
      & (n=1515) & (n=557) & (n=271) \\
      & Train Test & Train Test & Train Test \\
\hline
\abovespace
Log Reg		& 100\% 100\% & 100\% 100\% & 100\% 96.9\% \\
(Binary) \\    
Log Reg		& 100\% 94.9\% & 100\% 100\% & NA NA \\
(Softmax) \\
SVM			&  100\%  96\% & 100\% 71.5\% & 98.1\% 98.1\% \\
Mult Gauss 	&  94\%  92.7\% &  100\%  100\% &  98\%  85.4\% \\
(pass) \\
Mult Gauss &  NA  92.4\% &  NA  92.7\% &  NA  78.6\% \\
(fail) \\
\hline
\end{tabular}
\end{sc}
\end{small}
\end{center}
\end{table}

% Conclusion and future work
\section{Conclusion and Future Work}
In this paper, we successfully implemented and compared variations of logistic regression, SVM, and multiple Gaussian machine learning algorithms to predict whether a given smart phone user was the authorized user given a set of gestures.  We identified areas where our methodologies expanded upon recent experiments conducted by other teams and areas like Multiple Gaussians where our approaches are new and promising.  \newline
Two areas for future work include improved and expanded data collection approaches and further expansion into unsupervised learning methods.  Our experiments were limited by the quantity and repetitiveness of the data we collected.  We recommend expanded data collection across an expanded set of users with more samples per user collected in a more natural approach.  Capturing device gyroscopic data would enhance the data by including 3-D phone position features.  Improved data quality and increased data quantity would allow additional unsupervised methods including the implementation of a neural network to capture non-linear relationships in the data. \newline
We have seen the need for a reduced feature set, and only a small subset of latent variables contain most variability in data. Factor analysis can be used to find a reduced set of features that inherently dictate the behavior of a finger.  In this situation, we say there is a hidden latent variable called z \( {\in \mathbf{R}^m } \) where our feature set x \(	\in \mathbf{R}^n\) and m \(\le\) n. When we assume that \( z \sim \mathcal{N} (0,I) \) and \(	{x|z \sim \mathcal{N}(u + \Lambda z,\Psi)	}\),  we iteratively use EM to find the hidden latent variable and matrix \({\Lambda }\) that describes the behavior of our feature set. 
\newline
Overall, we are pleased with our results and hope our experiments help advance gesture based smart phone security.

\newpage
\newpage
\section{References}
[1] Feng, Tao, et al. "Tips: Context-aware implicit user identification using touch screen in uncontrolled environments." \textit{Proceedings of the 15th Workshop on Mobile Computing Systems and Applications.} ACM, 2014.\newline

[2] Meng, Yuxin, Duncan S. Wong, and Roman Schlegel. "Touch gestures based biometric authentication scheme for touchscreen mobile phones." \textit{International Conference on Information Security and Cryptology.} Springer, Berlin, Heidelberg, 2012. \newline

[3] Li, Lingjun, Xinxin Zhao, and Guoliang Xue. "Unobservable Re-authentication for Smartphones." \textit{NDSS}. 2013.\newline

[4] Zheng, Nan, et al. "You are how you touch: User verification on smartphones via tapping behaviors." \textit{Network Protocols (ICNP), 2014 IEEE 22nd International Conference on}. IEEE, 2014.
APA \newline

[5] Bo, Cheng, et al. "Continuous user identification via touch and movement behavioral biometrics." \textit{Performance Computing and Communications Conference (IPCCC), 2014 IEEE International}. IEEE, 2014. \newline

[6] Pedregosa, Fabian, et al. "Scikit-learn: Machine learning in Python." \textit{Journal of Machine Learning Research} 12.Oct (2011): 2825-2830.\newline

[7] MATLAB and Statistics and Machine Learning Toolbox Release 2017b, The MathWorks, Inc., Natick, Massachusetts, United States.\newline

% Contribitions
\newpage
\section{Contributions}

\begin{itemize}
\item John Peruzzi - Linear models, one class SVM(not used), poster creation, poster session, paper writing
\item Philip Andrew Wingard - Data collection, Data Filtering / Creation of Feature Sets, Multiple Gaussian Anomaly Detector, poster creation, paper writing
\item David Zucker - SVM modeling, Feature generation, Recursive model generation, visuals, poster creation, paper writing
\end{itemize}

Code Repository Link\newline
\url{https://drive.google.com/drive/folders/1rYS68xrafSDNUP63YAgNFDchiC7tI2hn?usp=sharing}

% Side by side figures 
%%\begin{figure*}
%\vskip 0.2in
%\begin{center}
%\begin{minipage}[c]{0.3\linewidth} 
%\includegraphics[width=\linewidth]{TAPS_5_SVM}
%\end{minipage}
%\hfill
%\begin{minipage}[c]{0.3\linewidth}
%\includegraphics[width=\linewidth]{Circle_30_SVM_Post_Reduction}
%\end{minipage}%
%\hfill
%v\begin{minipage}[c]{0.3\linewidth}
%\includegraphics[width=\linewidth]{Random_30_SVM_Post_FeatureAdjustment}
%\end{minipage}%
%\caption{SVM classification results across datasets using parameter grid %search for C and $\gamma$ between (1e-03 thru 1e+10) and (1e-15 thru v1e+03), respectively.}
%\end{center}
%\vskip -0.2in
%v\end{figure*}

\end{document}